\documentclass{article}
\usepackage{pdfpages}
\pdfoutput=1
\usepackage[preprint, nonatbib]{neurips_2019}
\usepackage{microtype}
\usepackage{graphicx}
\usepackage{color}
\usepackage{subfigure}
\usepackage{amsmath, amsthm, amssymb, float}
\usepackage{booktabs} 
\usepackage{tikz}
\usepackage{bm}
\usepackage{mathtools}
\usetikzlibrary{matrix,chains,positioning,decorations.pathreplacing,arrows}
\usetikzlibrary{bayesnet}
\usepackage{graphicx}

\newtheorem{theorem}{Theorem}
\usepackage{siunitx}
\usepackage{etoolbox}
\usepackage{algorithm}
\usepackage{algorithmic}

\newlength\figH
\newlength\figW
\usepackage[colorinlistoftodos]{todonotes}
\usepackage{pgfplots}  
\usepackage{mathtools}
\usetikzlibrary{positioning, arrows}
\usetikzlibrary{external}
\usetikzlibrary{plotmarks}
\usetikzlibrary{plotmarks}
\usepackage{grffile}
\pgfplotsset{compat=newest} 
\pgfplotsset{plot coordinates/math parser=false} 
\usetikzlibrary{arrows}
\usetikzlibrary{decorations.markings}
\usepackage{url}
\title{Solving graph compression via optimal transport}

\author{%
  Vikas K.~Garg\\
  MIT\\
  \texttt{vgarg@csail.mit.edu} \\
  \And
  Tommi Jaakkola \\
  MIT \\
  \texttt{tommi@csail.mit.edu}\\
  }

\usepackage[numbers]{natbib}
\begin{document}

\maketitle

\begin{abstract}
We propose a new approach to graph compression by appeal to optimal transport. The transport problem is seeded with prior information about node importance, attributes, and edges in the graph. The transport formulation can be setup for either directed or undirected graphs, and its dual characterization is cast in terms of distributions over the nodes. The compression pertains to the support of node distributions and makes the problem challenging to solve directly. To this end, we introduce Boolean relaxations and specify conditions under which these relaxations are exact. The relaxations admit algorithms with provably fast convergence. Moreover, we provide an exact $O(d \log d)$ algorithm for the subproblem of projecting a $d$-dimensional vector to transformed simplex constraints. Our method outperforms state-of-the-art compression methods on graph classification. 
\end{abstract}

\section{Introduction}
Graphs are widely used to capture complex relational objects, from social interactions to molecular structures. Large, richly connected graphs can be, however, computationally unwieldy if used as-is, and spurious features present in the graphs that are unrelated to the task can be statistically distracting. A significant effort thus has been spent on developing methods for compressing or summarizing graphs towards graph {\em sketches} \cite{LV2018}. Beyond computational gains, these sketches take center stage in numerous tasks pertaining to graphs such as partitioning  (\cite{DGK2007, WXSW2014}), unraveling complex or multi-resolution structures (\cite{RB2003, SD2011, KTG2014, TMK2016}), obtaining coarse-grained diffusion maps (\cite{LL2006}), including neural convolutions (\cite{BZSL2014, DBV2016, SK2017}). State-of-the-art compression methods broadly fall into two categories: (a) sparsification (removing edges)  and (b) coarsening (merging vertices).  These methods measure spectral similarity between the original graph and a compressed representation in terms of a (inverse) Laplacian quadratic form \cite{RSB2011, CS2011, LB2012, SFV2016, DB2013}. Thus, albeit these methods approximately preserve the graph spectrum, e.g., \cite{LV2018}; they are oblivious to, and thus less effective for, downstream tasks such as classification that rely on labels or attributes of the nodes. Also, the key compression steps in these methods are, typically, either heuristic or detached from their original objective \cite{HG2019}.  

We address these issues by taking a novel perspective that appeals to the theory of optimal transport \cite{V2008}, and develops its connections to minimum cost flow on graphs \cite{K2008, CMSV2017}. Specifically, we interpret graph compression as minimizing the transport cost from a fixed initial distribution supported on all vertices to an unknown target distribution whose size of support is limited by the amount of compression desired. Thus, the compressed graph in our case is a subgraph of the original graph, restricted to a subset of the vertices selected via the associated transport problem. The transport cost depends on the specified prior information such as importance of the nodes and their labels or attributes, and thus can be informed by the downstream task. Moreover, we take into account the underlying geometry toward the transport cost, unlike agnostic measures such as KL-divergence \cite{SRGB2016}.


There are several technical challenges that we must address. First, the standard notion of optimal transport on graphs is tailored to directed graphs where the transport cost decomposes as a directed flow along the edge orientations (\cite{ES2018}). To circumvent this limitation, we extend optimal transport on graphs to handle both directed and undirected edges, and derive a dual that directly measures the discrepancy between distributions on the vertices. As a result, we can also compress {\em mixed } graphs \cite{HKW1997, R2007, CDLS2007, BBCJY2015}.
The second challenge comes from the compression itself, enforced in our approach as sparse support of the target distribution.  Optimal transport is known to be computationally intensive, and almost all recent applications of OT in machine learning, e.g., \cite{ACB2017, CFHR2017, RCFT2019, FZMAP2015, AJJ2018, VCFTC2018} rely on entropy regularization \cite{C2013, BCCNP2015, S2015, GCPB2016, SDFCRB2018} for tractability.
However, entropy is not conducive to sparse solutions since it discourages the variables from ever becoming 0 \cite{ES2018}. In principle, one could consider convex alternatives such as enforcing $\ell_1$ penalty (e.g., \cite{T1996}). However, such methods require iterative tuning to find a solution that matches the desired support, and  require strong assumptions such as restricted eigenvalue, isometry, or nullspace conditions for recovery. Some of these issues were previously averted by introducing binary selection variables \cite{TTW2014, PWG2015} and using Boolean relaxations in the context of unconstrained real spaces (regression). However, they do not apply to our setting since the target distribution must reside in the simplex. We introduce {\em constrained} Boolean relaxations that are not only efficient, but also provide exactness certificates.

Our graph compression formulation also introduces new algorithmic challenges. For example, solving our sparsity controlled dual transport problem involves a new subproblem of projecting on the probability simplex $\Delta$ under a diagonal transformation. Specifically, let $D(\epsilon)$ be a diagonal matrix with the diagonal $\epsilon \in [0, 1]^d \setminus \{\boldsymbol{0}\}$. Then, for a given $\epsilon$, the problem is to find the projection $x \in \mathbb{R}^d$ of a given vector $y \in  \mathbb{R}^d$ such that $D(\epsilon) x \in \Delta$. This generalizes well-studied problem of Euclidean projection on the probability simplex (\cite{DSSC2008,WC2013, C2016}), recovered if each $\epsilon_i$ is set to 1. We provide an exact $O(d \log d)$ algorithm for solving this generalized projection. Our approach leads to convex-concave saddle point problems with fast convergence via methods such as Mirror Prox \cite{N2004}. 

\textbf{Our contributions.} We propose an approach for graph compression based on optimal transport (OT). Specifically, we (1) extend OT to undirected and mixed graphs (section \ref{OTGeneral}), (2) introduce constrained Boolean relaxations for our dual OT problem, and provide exactness guarantees (section \ref{ConBool}) (3) generalize Euclidean projection onto simplex, and provide an efficient algorithm (section \ref{ConBool}), and (4) demonstrate that our algorithm outperforms state-of-the-art  
compression methods, both in accuracy and compression time, on classifying graphs from standard real datasets. We also provide qualitative results that our approach provides meaningful compression in synthetic and real graphs (section \ref{Experiments}).

\section{Optimal transport for general edges} \label{OTGeneral}
Let $\vec{G} = (V, \vec{E})$ be a directed graph on nodes (or vertices) $V$ and edges $\vec{E}$. We define the signed incidence matrix $\vec{F}$: $\vec{F}(\vec{e}, v) = 1$ if $\vec{e} = (w, v) \in \vec{E} \mbox{ for some } w \in V$, $-1$ if $\vec{e} = (v, w) \in \vec{E} \mbox{ for some } w \in V$, and $0$ otherwise.
Let $c(\vec{e}) \in \mathbb{R}_+$ be the positive cost to transport unit mass along edge $\vec{e} \in \vec{E}$, and $\Delta(V)$ the probability simplex on $V$. The shorthand $a \preceq b$ denotes that $a(i) \leq b(i)$ for each component $i$. Let $\boldsymbol{0}$ be a vector of all zeros and $\boldsymbol{1}$ a vector of all ones.
Let $\rho_0, \rho_1 \in \Delta(V)$ be  distributions over the vertices in $V$. The optimal transport distance $\vec{W}(\rho_0, \rho_1)$ from $\rho_0$ to $\rho_1$ is \cite{ES2018}:
\begin{eqnarray*} 
\vec{W}(\rho_0, \rho_1) & = & \displaystyle \min_{\substack{J \in \mathbb{R}^{|\vec{E}|}\\~~ \boldsymbol{0} ~\preceq~ J}}  ~~ \sum_{\vec{e} \in \vec{E}} c(\vec{e}) J(\vec{e})~~ \qquad \text{s.t.} ~~~~~\vec{F}^{\top} J  ~=~ \rho_1 - \rho_0~,
\end{eqnarray*}
where $J(\vec{e})$ is the non-negative mass transfer from tail to head on edge $\vec{e}$. Intuitively, $\vec{W}(\rho_0, \rho_1)$ is the minimum cost of a directed flow from $\rho_0$ to $\rho_1$.  In order to extend this intuition to the undirected graphs, we need to refine the notion of incidence, and let the mass flow in either direction. 
Specifically, let $G = (V, E)$ be a connected undirected graph.  We define the incidence matrix pertaining to $G$ as $F(e, v) = 1  \mbox { if  edge } e  \text{ is incident on } v$, and $0$ otherwise.
 \noindent With each undirected edge $e \in E$, having cost $c(e) \in \mathbb{R}_+$, we associate two directed edges $e^+$ and $e^-$, each with cost $c(e)$, and flow variables $J^+(e), J^-(e) \geq 0$. Then, the total {\em undirected flow} pertaining to $e$ is $J^+(e) + J^-(e)$. 
 Since we incur cost for flow in either direction, we define the optimal transport cost $W(\rho_0, \rho_1)$ from $\rho_0$ to $\rho_1$ as  
\begin{eqnarray} 
 \min_{\substack{J^+, J^- \in \mathbb{R}^{|E|} \\ \boldsymbol{0} ~\preceq~ J^+, J^-}} & \displaystyle \sum_{e \in E} c(e) (J^+(e) + J^-(e)) \qquad
\text{s.t.} \quad F^{\top} (J^- - J^+)  =  \rho_1 - \rho_0~~.  \label{UnWass}
\end{eqnarray}
We call a directed edge $e^+$ {\em active} if $J^+(e) > 0$, i.e., there is some positive flow on the edge (likewise for $e^-$). Moreover, by extension, we call an undirected edge $e$ active if at least one of $e^+$ and $e^-$ is active. We claim that at most one of $e^+$ and $e^-$ may be active for any edge $e$.
\begin{theorem}\label{Theorem1}
The optimal solution to \eqref{UnWass} must have  $J^+(e) = 0$ or $J^-(e) = 0$ (or both) $\forall\,e \in E$.   
\end{theorem}
The proof is provided in the supplementary material. Thus, like the directed graph setting, we either have flow in only one direction for each edge $e$, or no flow at all. Moreover, Theorem \ref{Theorem1} facilitates generalizing optimal transport distance to mixed graphs $\tilde{G}(V, E, \vec{E})$, i.e., where both directed and undirected edges may be present. In particular, we adapt the formulation in \eqref{UnWass} with minor changes: (a) we associate bidirectional variables with each edge, directed or undirected. For the undirected edges $e \in E$, we replicate the constraints from \eqref{UnWass}. For the directed edges $\vec{e}$, we  follow the convention that $J^{+}(\vec{e})$ denotes the outward flow along $\vec{e}$ whereas $J^{-}(\vec{e})$ denotes the incoming flow (from head to tail), and impose the additional constraints $J^{-}(\vec{e}) = 0$. We will focus on undirected graphs $G = (V, E)$ since the extensions to directed and mixed graphs are immediate due to Theorem \ref{Theorem1}.

\section{Graph compression} \label{SecGraphCompress}
We view graph compression as the problem of minimizing the optimal transport distance from an initial distribution $\rho_0$ having full support on the vertices $V$ to a target distribution $\rho_1$ that is supported only on a subset $S_V(\rho_1)$ of $V$. The compressed subgraph is obtained by restricting the original graph to vertices in $S_V(\rho_1)$ and the incident edges. The initial distribution $\rho_0$ encodes any prior information. For instance, it might be taken as a stationary distribution of random walk on the graph. Likewise, the cost function $c$ encodes the preference for different edges. In particular, a high value of $c(e)$ would inhibit edge $e$ from being active. This flexibility allows our framework to inform compression based on the specifics of different downstream applications by getting to define $\rho_0$ and $c$ appropriately.  

\subsection{Dual characterization of the transport distance} \label{DualChar}
Note that \eqref{UnWass} defines an optimization problem over edges. However, our perspective requires quantifying $W(\rho_0, \rho_1)$ as an optimization over the vertices. Fortunately, strong duality comes to our rescue. Let $c = (c(e), e \in E)$ be the column vector obtained by stacking the costs. The dual of \eqref{UnWass} is 
\begin{eqnarray} 
 \max_{\substack{\boldsymbol{0} \preceq y, \boldsymbol{0} \preceq z \\ - c ~\preceq~ F (y - z) \preceq  c}} & (y-z)^{\top}(\rho_1 - \rho_0),~~~~ \text{or equivalently},~~~  \displaystyle \max_{\substack{t \in \mathbb{R}^{|V|}\\ - c   ~\preceq Ft  \preceq  c}} ~~ t^{\top}(\rho_1 - \rho_0)~~. \label{UnWassDual}
 \end{eqnarray}
This alternative formulation of $W(\rho_0, \rho_1)$ in \eqref{UnWassDual} lets us define compression solely in terms of variables over vertices. Specifically, for a budget of at most $k$ vertices, we solve 
\begin{eqnarray} \label{CompressOriginal}
\min_{\substack{\rho_1 \in \Delta(V)\\ ||\rho_1||_0 \leq k}}~~~ \max_{\substack{t \in \mathbb{R}^{|V|}\\ - c \preceq Ft  \preceq c}} \underbrace{t^{\top}(\rho_1 - \rho_0) + \dfrac{\lambda}{2} ||\rho_1||^2}_{\mathcal{L}_\lambda(\rho_1, t;\rho_0)} ~,&& 
\end{eqnarray}
where $\lambda > 0$ is a regularization hyperparameter, and  $||\rho_1||_0$ is the number of vertices with positive mass under $\rho_1$, i.e., the cardinality of support set $S_V(\rho_1)$. The quadratic penalty is strongly convex in $\rho_1$, so as we shall see shortly, would help us leverage fast algorithms for a saddle point problem. 
Note that a high value of $\lambda$ would encourage $\rho_1$ toward a uniform distribution. We favor this penalty over entropy, which is not conducive to sparse solutions since entropy would forbid $\rho_1$ from having zero mass at any vertex in the graph.  

Our next result reveals the structure of optimal solution $\rho_1^*$ in \eqref{CompressOriginal}. Specifically, $\rho_1^*$ must be expressible as an affine function of $\rho_0$ and $F$. Moreover, the constraints on active edges are tight. This reaffirms our intuition that $\rho_1^*$ is obtained  from $\rho_0$ by transporting mass along a subset of the edges, i.e., the active edges. The remaining edges do not participate in the flow.  

\begin{theorem} \label{Theorem2}
The optimal $\rho_1^*$ in \eqref{CompressOriginal} is of the form $\rho_1^* = \rho_0 + F^{\top}\eta~,$ where $\eta \in \mathbb{R}^{|E|}$.  Furthermore, for any active edge $e \in E$, we must have $Ft^*(e) \in \{c(e), - c(e)\}$.  
\end{theorem}

 
\subsection{Constrained Boolean relaxations} \label{ConBool}
 The formulation \eqref{CompressOriginal} is non-convex due to the support constraint on $\rho_1$. Since recovery under $\ell_1$ based methods such as Lasso often requires extensive tuning, we resort to the method of Boolean relaxations that affords an explicit control much like the $\ell_0$ penalty. However, prior literature on Boolean relaxations is limited to variables that have no additional constraints beyond sparsity. Thus, in order to deal with the simplex constraints $\rho_1 \in \Delta(V)$, we introduce constrained Boolean relaxations. Specifically, we define the characteristic function $g_V(x) = 0 \mbox{ if }  x \in \Delta(V)$ and $\infty$ otherwise, 
and move the non-sparsity constraints inside the objective. This lets us delegate the sparsity constraints to binary variables, which can be relaxed to $[0, 1]$. Using the definition of 
$\mathcal{L}_\lambda$, we can write $\eqref{CompressOriginal}$ as
$$\min_{\substack{\rho_1 \in \mathbb{R}^{|V|}\\ ||\rho_1||_0 \leq k}}~~~ \max_{\substack{t \in \mathbb{R}^{|V|}\\ - c \preceq Ft  \preceq c}} \mathcal{L}_\lambda(\rho_1, t;\rho_0) ~+~ g_V(\rho_1) ~.$$
Denoting by $\odot$ the Hadamard (elementwise) product, and introducing  variables $\epsilon \in \{0, 1\}^{|V|}$, we get
\begin{eqnarray*} 
\min_{\substack{\epsilon \in \{0, 1\}^{|V|}\\ ||\epsilon||_0 \leq k}} \min_{\rho_1 \in \mathbb{R}^{|V|}} ~~~ \max_{\substack{t \in \mathbb{R}^{|V|}\\ - c \preceq Ft  \preceq c}} \mathcal{L}_\lambda(\rho_1 \odot \epsilon, t;\rho_0) ~+~ g_V(\rho_1 \odot \epsilon) ~. 
\end{eqnarray*}
Adjusting the characteristic term as a constraint, we have the following equivalent problem \begin{eqnarray} \label{EqForProj}
\min_{\substack{\epsilon \in \{0, 1\}^{|V|}\\ ||\epsilon||_0 \leq k}}~~~ \min_{\substack{\rho_1 \in \mathbb{R}^{|V|} \\ \rho_1 \odot \epsilon \in \Delta(V)}} \max_{\substack{t \in \mathbb{R}^{|V|}\\ - c \preceq Ft  \preceq c}} \mathcal{L}_\lambda(\rho_1 \odot \epsilon, t;\rho_0) ~. 
\end{eqnarray}  
\begin{minipage}[p]{0.45\textwidth}
\begin{algorithm}[H]
\caption{Algorithm  to compute \hspace{3cm} \mbox{~~} Euclidean projection on the $d$-simplex $\Delta$ under a diagonal transformation.}
\label{ProjectWS}
\begin{algorithmic}
\STATE {\bfseries Input:} $y$, $\epsilon$
\vspace{0.5ex}
\STATE Define $\mathcal{I}_{>}  \triangleq \{j \in [d] ~|~ \epsilon_j > 0\}$ 
\STATE Define $\mathcal{I}_{=} \triangleq \{j \in [d] ~|~ \epsilon_j = 0\}$;
\STATE $y_{>}  \triangleq \{y_j ~|~ j \in \mathcal{I}_{>}\}$; $\epsilon_{>}  \triangleq \{\epsilon_j ~|~ j \in \mathcal{I}_{>}\}$
\STATE Sort $y_>$ into $\hat{y}_>$ and $\epsilon_>$ into $\hat{\epsilon}_>$, in non-increasing order, based on  $y_{j}/\epsilon_{j}, j \in \mathcal{I}_{>}$~. Rename indices in ($\hat{y}_>$, $\hat{\epsilon}_>$) to start from 1. Let $\pi$ map $j \in \mathcal{I}_{>}$ to $\pi(j) \in [|\hat{y}_>|]$. Thus 
$$\hat{y}_1/\hat{\epsilon}_1 \geq \hat{y}_2/\hat{\epsilon}_2 \geq \ldots \geq \hat{y}_{|\mathcal{I}_{>} |}/\hat{\epsilon}_{|\mathcal{I}_{>} |}$$
\STATE $b_j =  \hat{y}_j + \hat{\epsilon}_j \dfrac{(1 - \sum_{i=1}^j \hat{\epsilon}_i \hat{y}_i)}{\sum_{i=1}^j \hat{\epsilon}_i^2}, ~ \forall j \in [| y_{>} |]$
\STATE $\ell ~=~ \max\left\{j \in [|y_{>}|] ~|~ b_j > 0\right\}$
\STATE $\alpha ~=~ \dfrac{(1 - \sum_{i=1}^{\ell} \hat{\epsilon}_i\hat{y}_i)}{\sum_{i=1}^\ell \hat{\epsilon}_i^2}$ \\
\STATE $x_j = \max\{\hat{y}_{\pi(j)} + \alpha \hat{\epsilon}_{\pi(j)}, 0\},~ \forall\,j \in  \mathcal{I}_{>}$
\STATE $x_j = y_{j}, \forall\,j \in  \mathcal{I}_{=}$
\end{algorithmic}
\end{algorithm} 
\end{minipage}\hfill
\begin{minipage}[p]{0.54\textwidth}
\begin{algorithm}[H]
\caption{Mirror Prox algorithm to (approximately) find $\epsilon$ in relaxation of \eqref{Dual2}. The step-sizes at time $\ell$ with respect to $\epsilon$, $t$, and $\zeta$ are $\alpha_{\ell}$, $\beta_{\ell}$, and $\gamma_{\ell}$ respectively.}
\label{MPPrimal1}
\begin{algorithmic}
\STATE {\bfseries Input:}  $\rho_0$,  $k$, $\lambda$; iterations $T$  
\STATE Define $\tilde{\mathcal{E}}_{k} =  \{\epsilon \in [0, 1]^{|V|} ~|~ \epsilon^{\top}\boldsymbol{1} \leq k\}$
\STATE Define $\mathcal{T}_{F, c}$  as in \eqref{TFCformula}
\STATE Define $\psi_{\rho_0}(\epsilon, t, \zeta)$ as in \eqref{Dual2}
\STATE Initialize $\epsilon^{(0)} = k\boldsymbol{1}/|V|$, $t^{(0)} = \boldsymbol{0}$, and $\zeta^{(0)}~=~0$
\FOR{$\ell = 0, 1,  \ldots, T$}
\STATE \quad~\underline{Gradient step:}
\vspace{0.5ex}
\STATE $\hat{\epsilon}^{(\ell)} = {\rm Proj}_{\tilde{\mathcal{E}}_{k}}\bigl(\epsilon^{(\ell)}-\alpha_{\ell} \nabla_{\epsilon} \psi_{\rho_0}(\epsilon^{(\ell)}, t^{(\ell)}, \zeta^{(\ell)})\bigr)$
\STATE $\hat{t}^{(\ell)}\!= {\rm Proj}_{\mathcal{T}_{F, c}}\bigl(
t^{(\ell)}\!+\beta_{\ell} \nabla_{t}\psi_{\rho_0}(\epsilon^{(\ell)}, t^{(\ell)}, \zeta^{(\ell)})\bigr)$ 
\STATE $\hat{\zeta}^{(\ell)}\!= 
\zeta^{(\ell)}\!+\gamma_{\ell} \nabla_{\zeta}\psi_{\rho_0}(\epsilon^{(\ell)}, t^{(\ell)}, \zeta^{(\ell)})$
\vspace{0.5ex}
\STATE {\quad~\underline{Extra-gradient step:}} 
\vspace{0.5ex}
\STATE $\epsilon^{(\ell+1)} = {\rm Proj}_{\tilde{\mathcal{E}}_{k}}\bigl(\epsilon^{(\ell)}-\alpha_{\ell} \nabla_{\epsilon} \psi_{\rho_0}(\hat{\epsilon}^{(\ell)}, \hat{t}^{(\ell)}, \hat{\zeta}^{(\ell)})\bigr)$
\STATE $t^{(\ell+1)}\!\!= \mathrm{Proj}_{\mathcal{T}_{F, c}}\bigl(
t^{(\ell)}\!+\beta_{\ell} \nabla_{t}\psi_{\rho_0}(\hat{\epsilon}^{(\ell)}, \hat{t}^{(\ell)}, \hat{\zeta}^{(\ell)})\bigr)$ 
\STATE $\zeta^{(\ell+1)}= 
\zeta^{(\ell)}\!+\gamma_{\ell} \nabla_{\zeta}\psi_{\rho_0}(\hat{\epsilon}^{(\ell)}, \hat{t}^{(\ell)}, \hat{\zeta}^{(\ell)})$
\ENDFOR
\vspace{0.5ex}
\STATE $\hat{\epsilon}=\sum_{\ell=1}^T\alpha_{\ell}\,\hat{\epsilon}^{(\ell)}\big/\sum_{\ell=1}^T\alpha_{\ell}$
\normalsize
\end{algorithmic}
\end{algorithm}  
\end{minipage}

\noindent Our formulation in \eqref{EqForProj} requires solving a new subproblem, namely, Euclidean projection on the $d$-simplex $\Delta$ under a diagonal transformation. Specifically, let $D(\epsilon)$ be a diagonal matrix with the diagonal $\epsilon \in [0, 1]^d \setminus \{\boldsymbol{0}\}$. Then, for a given $\epsilon$, the problem is to find the projection $x \in \mathbb{R}^d$ of a given vector $y \in  \mathbb{R}^d$ such that $D(\epsilon) x = x \odot \epsilon \in \Delta$. This problem generalizes  Euclidean projection on the probability simplex (\cite{DSSC2008,WC2013, C2016}), which is recovered when we set $\epsilon$ to $\boldsymbol{1}$, i.e., an all-ones vector.  
Our next result shows that Algorithm \ref{ProjectWS} solves this problem exactly in $O(d \log d)$ time.

 \begin{theorem} \label{Theorem6}
Let $\epsilon \in [0, 1]^d \setminus \{\boldsymbol{0}\}$ be a given vector of weights, and $y \in \mathbb{R}^d$ be a given vector of values. Algorithm \ref{ProjectWS} solves the following problem in  $O(d \log d)$ time 
\begin{eqnarray*}
\min_{x \in \mathbb{R}^d ~:~ x \odot \epsilon \in \Delta} & \dfrac{1}{2} ||x - y||^2 ~.
\end{eqnarray*}
\end{theorem}

Theorem \ref{Theorem6} allows us to relax the problem \eqref{EqForProj}, and solve the relaxed problem efficiently since the projection steps on the other constraint sets can be solved by known methods \cite{BT2014, GDX2018}. 
Specifically, since $\epsilon$ consists of only zeros and ones, $||\epsilon||_0 =  ||\epsilon||_1 = \epsilon^{\top}\boldsymbol{1}$ and $\epsilon \odot \epsilon = \epsilon$. So, we can write \eqref{EqForProj} as
\begin{eqnarray*} 
\min_{\substack{\epsilon \in \mathcal{E}_{k}}}~~~ \min_{\substack{\rho_1 \in \mathbb{R}^{|V|} \\ \rho_1 \odot \epsilon \in \Delta(V)}} \max_{t \in \mathcal{T}_{F, c}} \mathcal{L}_\lambda(\rho_1 \odot \epsilon \odot \epsilon, t;\rho_0) ~,
\end{eqnarray*}  
where we denote the constraints for $\epsilon$ and $t$ respectively by \begin{eqnarray} \label{TFCformula} \mathcal{E}_{k} & \triangleq  \{\epsilon \in \{0, 1\}^{|V|} ~|~ \epsilon^{\top}\boldsymbol{1} \leq k\}~,~~~ \text{and} ~~~ \mathcal{T}_{F, c} & \triangleq  \{t \in \mathbb{R}^{|V|} ~|~ - c \preceq Ft  \preceq c \}~.
\end{eqnarray}
We can thus eliminate $\epsilon$ from the regularization term via a change of variable $\rho_1 \odot \epsilon \to \tilde{\rho}_1$  
\begin{eqnarray} \label{EqP1_past}
\min_{\substack{\epsilon \in \mathcal{E}_{k}}}~~~ \min_{\substack{\tilde{\rho}_1 \in \mathbb{R}^{|V|} \\ \tilde{\rho}_1 \odot \epsilon \in \Delta(V)}} \max_{t \in \mathcal{T}_{F, c}}~~ \mathcal{L}_0(\tilde{\rho}_1 \odot \epsilon, t;\rho_0) + \dfrac{\lambda}{2} ||\tilde{\rho}_1||^2~~. 
\end{eqnarray}


We note that \eqref{EqP1_past} is a mixed-integer program due to constraints $\mathcal{E}_{k}$, and thus hard to solve. Nonetheless, we can relax the hard binary constraints on the coordinates of $\epsilon$ to  $[0, 1]$ intervals to obtain a saddle point formulation with a strongly convex term, and solve the relaxation efficiently, e.g., via customized versions of methods such as Mirror Prox \cite{N2004}, Accelerated Gradient Descent \cite{N2005}, or Primal-Dual Hybrid Gradient \cite{CP2011}.   

An attractive property of our relaxation is that if the solution $\hat{\epsilon}$ from the relaxed problem is integral then $\hat{\epsilon}$ must be optimal for the non-relaxed hard problem \eqref{EqP1_past}, and so the original formulation \eqref{CompressOriginal}. We now pin down the  necessary and sufficient conditions for optimality of $\hat{\epsilon}$. 
\begin{theorem} \label{RelaxationExactness}
Let $S_V({\rho_1^*}) = \{v \in V ~|~ \rho_{1}^* (v) > 0 \}$ be the support of optimal $\rho_1$ in the original formulation \eqref{CompressOriginal}. Let the indicator $I_{S_V^*} \in \{0, 1\}^{|V|}$ be such that $I_{S_V^*}(v) = 1$ if $v \in S_V(\rho_1^*)$ and 0 otherwise.   
The relaxation of \eqref{EqP1_past} is guaranteed to recover $S_V({\rho_1^*})$   if and only if there exists a tuple $(\gamma, \hat{t}, \hat{\nu}, \hat{\zeta}) \in \mathbb{R}_+ \times \mathbb{R}^{|V|} \times \mathbb{R}_{+}^{|V|} \times \mathbb{R}$ such that the following holds for all vertices $v \in V$,
\begin{equation} \label{RelaxOptConditions}
|\hat{t}(v) - \hat{\nu}(v) + \hat{\zeta}| ~~~~ \begin{cases} >~ \gamma \qquad {\rm if }~~ v \in S_V(\rho_1^*)\\ <~ \gamma \qquad {\rm if }~~ v \notin S_V(\rho_1^*) \end{cases}, ~~~ \text{where}  
\end{equation}
\begin{eqnarray} \label{Opthats}
 (\hat{t}, \hat{\nu}, \hat{\zeta}) ~\in~ \arg \max_{t \in \mathcal{T}_{F, c}} \max_{\nu \in \mathbb{R}_+^{|V|}} \max_{\zeta \in \mathbb{R}} ~~- \left(~\dfrac{1}{2\lambda} \big|\big|(t - \nu + \zeta \boldsymbol{1})\odot I_{S_V^*}\big|\big|^2 ~+~ t^{\top} \rho_0 ~+~ \zeta \right)~.
\end{eqnarray} 
\end{theorem}
For some applications projecting on the simplex, as required by \eqref{EqP1_past}, may be an expensive operation. We can invoke the minimax theorem to swap the order of $\tilde{\rho}_1$ and $t$,  and proceed with a Lagrangian dual to eliminate $\tilde{\rho}_1$ at the expense of introducing a scalar variable. Thus, effectively, we can replace the projection on simplex by a one-dimensional search. We state this 
equivalent formulation below.  

\begin{theorem} \label{DualForm}
Problem \eqref{EqP1_past}, and thus the original formulation \eqref{CompressOriginal}, is equivalent to 
 \begin{eqnarray} \label{Dual2}
\displaystyle \min_{\epsilon \in  \mathcal{E}_{k}}~\max_{\substack{t \in \mathcal{T}_{F, c}\\\zeta \in \mathbb{R}}}   \underbrace{- \dfrac{1}{2\lambda} \sum_{v: t(v) \leq -\zeta} \left(\epsilon(v) (t(v) + \zeta)^2 ~+~ 2 \lambda t(v) \rho_{0}(v) \right) ~-~ \sum_{v: t(v) > -\zeta} t(v) \rho_{0}(v) -  \zeta}_{\psi_{\rho_0}(\epsilon, t, \zeta)}~.
\end{eqnarray}
\end{theorem}

We present a customized Mirror Prox procedure in Algorithm \ref{MPPrimal1}. The  projections ${\rm Proj}_{\mathcal{T}_{F, c}}$ and ${\rm Proj}_{\tilde{\mathcal{E}}_{k}}$  can be computed efficiently, respectively, by \cite{BT2014} and \cite{GDX2018}.  We round the solution $\hat{\epsilon} \in [0, 1]^{|V|}$ returned by the algorithm to have at most $k$ vertices as the estimated support for the target distribution $\rho_1$ if $\hat{\epsilon}$ is not integral.  The compressed graph is taken to be the subgraph spanned by these vertices.

\begin{table*}[t!]  \caption{{\bf Description of graph datasets, and comparison of accuracy on test data.}  We provide the statistics on the number of graphs, number of classes, average number of nodes, and average number of edges in each dataset. The classification test accuracy (along with standard deviation) when each graph was (roughly) compressed to half is shown for each method for each training fraction in $\{0.2, \ldots, 0.8\}$.  The algorithm having the best performance is indicated with bold font in each case. '-' entries indicate that the method failed to compress the dataset (e.g. due to matrix singularity).  
  \label{tab:compare}}   
  \vskip 0.1in 
  \addtolength{\tabcolsep}{-4.5pt} 
 \centering  
 \small 
  \begin{tabular} {|c|c|c|c|c|c|c|c|c|c|} 
  \toprule 
   Dataset &  method &  acc@0.2 &  acc@0.3 & acc@0.4 &  acc@0.5 &  acc@0.6 &  acc@0.7 & acc@0.8 \\   \hline 
 \toprule 
{\bf MSRC-21C} & REC & .485$\pm$.016 & .543$\pm$.010 & .595$\pm$.010 & .625$\pm$.013 & .641$\pm$.008 & .696$\pm$.013 & .738$\pm$.016\\
graphs: 209 & Heavy & .408$\pm$.016 & .479$\pm$.015 & .516$\pm$.009 & .538$\pm$.009 & .557$\pm$.011 & .602$\pm$.022 & .653$\pm$.011\\
classes: 20 & Affinity & .413$\pm$.021 & .489$\pm$.008 & .516$\pm$.011 & .549$\pm$.010 & .560$\pm$.016 & .607$\pm$.019 & .654$\pm$.021\\
nodes: 40.3 & Alg. Dist. & .452$\pm$.036 & .498$\pm$.035 & .524$\pm$.021 & .535$\pm$.027 & .531$\pm$.029 & .590$\pm$.032 & .652$\pm$.044\\
edges: 96.6 & OTC & {\bf .548$\pm$.004} & {\bf .605$\pm$.003} & {\bf .639$\pm$.006} & {\bf .679$\pm$.003} & {\bf .696$\pm$.002} & {\bf .742$\pm$.007} & {\bf .778$\pm$.005}\\
\toprule 
{\bf DHFR} & REC & .681$\pm$.011 & .704$\pm$.014 & .724$\pm$.007 & .738$\pm$.009 & .749$\pm$.008 & .756$\pm$.011 & .771$\pm$.011\\
graphs: 467 & Heavy & .719$\pm$.010 & .751$\pm$.010 & .776$\pm$.012 & .782$\pm$.008 & .777$\pm$.009 & .786$\pm$.014 & .799$\pm$.013\\
classes: 2 & Affinity & .717$\pm$.013 & .733$\pm$.011 & .745$\pm$.014 & .761$\pm$.014 & .771$\pm$.019 & .767$\pm$.015 & .785$\pm$.013\\
nodes: 42.4 & Alg. Dist. & .743$\pm$.011 & .761$\pm$.012 & .768$\pm$.022 & .786$\pm$.019 & .810$\pm$.025 & \bf .817$\pm$.033 & .809$\pm$.030\\
edges: 44.5 & OTC & {\bf .757$\pm$.004} & {\bf .784$\pm$.003} & {\bf .797$\pm$.005} & {\bf .799$\pm$.003} & {\bf .811$\pm$.007} & .814$\pm$.006 & {\bf .823$\pm$.004}\\
\toprule
{\bf MSRC-9} & REC & .738$\pm$.011 & .782$\pm$.010 & .817$\pm$.009 & .818$\pm$.013 & .835$\pm$.020 & .833$\pm$.018 & .840$\pm$.013\\
graphs: 221 & Heavy & .648$\pm$.019 & .710$\pm$.024 & .766$\pm$.014 & .773$\pm$.010 & .786$\pm$.009 & .796$\pm$.010 & .813$\pm$.009\\
classes: 8 & Affinity & .665$\pm$.015 & .722$\pm$.005 & .762$\pm$.010 & .774$\pm$.014 & .789$\pm$.026 & .786$\pm$.019 & .801$\pm$.017\\
nodes: 40.6 & Alg. Dist. & .666$\pm$.048 & .717$\pm$.051 & .756$\pm$.029 & .771$\pm$.039 & .798$\pm$.032 & .803$\pm$.030 & .809$\pm$.046\\
edges: 97.9 & OTC & {\bf .784$\pm$.005} & {\bf .808$\pm$.005} & {\bf .826$\pm$.007} & {\bf .846$\pm$.003} & {\bf .839$\pm$.006} & {\bf .842$\pm$.007} & {\bf .854$\pm$.003}\\
\toprule 
{\bf BZR-MD} & REC & .525$\pm$.011 & .548$\pm$.015 & {\bf .563$\pm$.020} & .553$\pm$.021 & .563$\pm$.012 & .569$\pm$.012 & .587$\pm$.020\\
graphs: 306 & Heavy & .497$\pm$.000 & .546$\pm$.000 & .555$\pm$.000 & .522$\pm$.000 & .550$\pm$.000 & .572$\pm$.000 & .558$\pm$.000\\
classes: 2 & Affinity & .508$\pm$.006 & .534$\pm$.012 & .534$\pm$.017 & .532$\pm$.015 & .549$\pm$.020 & .567$\pm$.033 & .562$\pm$.029\\
nodes: 21.3 & Alg. Dist. & .497$\pm$.021 & .546$\pm$.026 & .555$\pm$.038 & .522$\pm$.028 & .550$\pm$.028 & .572$\pm$.024 & .558$\pm$.039\\
edges: 225.06 & OTC & {\bf .534$\pm$.000} & {\bf .569$\pm$.000} &  .547$\pm$.000 & {\bf .579$\pm$.000} & {\bf .572$\pm$.000} & {\bf .607$\pm$.000} & {\bf .603$\pm$.000}\\
\toprule 
{\bf Mutagenicity} & REC & .713$\pm$.006 & .730$\pm$.006 & .742$\pm$.005 & .752$\pm$.004 & .758$\pm$.005 & .765$\pm$.007 & .769$\pm$.007\\
graphs: 4337 & Heavy & .718 $\pm$.006 & .738$\pm$.004 & .753$\pm$.004 & .763$\pm$.004 & .771$\pm$.003 & .779$\pm$.004 & .783$\pm$.004\\
classes: 2 & Affinity & - & - & - & - & - & - & -\\
nodes: 30.3 & Algebraic & - & - & - & - & - & - & -\\
edges: 30.8 & OTC & {\bf .749$\pm$.002} & {\bf .768$\pm$.003} & {\bf .779$\pm$.003} & {\bf .787$\pm$.004} & {\bf .792$\pm$.004} & {\bf .795$\pm$.003} & {\bf .799$\pm$.003}\\
\toprule 
 \end{tabular} 
  \label{default} 
  \end{table*}

\section{Experiments} \label{Experiments}
We conducted several experiments to demonstrate the merits of our method. 
We start by describing the experimental setup. We fixed the value of hyperparameters in Algorithm \ref{MPPrimal1} for all our experiments. Specifically, we set the regularization coefficient $\lambda=1$, and the gradient rates $\alpha_\ell=0.1, \beta_\ell=0.1, \gamma_\ell=0.1$ for each $\ell \in \{0, 1, \ldots, T\}$. We also let $\rho_0$ be the stationary distribution by setting $\rho_0(v)$ for each $v \in V$ as the ratio of $deg(v)$, i.e. the degree of $v$, to the sum of degrees of all the vertices. Note that the distribution thus obtained is the unique stationary distribution for connected non-bipartite graphs, and {\em a} stationary distribution for the bipartite graphs. Moreover, for non-bipartite graphs, it has a nice physical interpretation in that any random walk on the graph always converges to this distribution irrespective of the graph structure. 

The objective of our experiments is three-fold. Since compression is often employed as a preprocessing step for further tasks, we first show that our method compares favorably, in terms of test accuracy, to the state-of-the-art compression methods on graph classification. We then demonstrate that our method performed the best in terms of the compression time. We finally show that our approach provides qualitatively meaningful compression on synthetic and real examples.     

\subsection{Classifying standard graph data} \label{ClassifySection}
We used several standard graph datasets for our experiments, namely, DHFR \cite{SOW2003}, BZR-MD \cite{KM2012}, MSRC-9, MSRC-21C \cite{NGBK2016}, and  Mutagenicity \cite{KMB2005}. We focused on these datasets since they represent a wide spectrum in terms of the number of graphs, number of classes, average number of nodes per graph, and average number of edges per graph (see Table \ref{tab:compare} for details). All these datasets have a class label for each graph, and additionally, labels for each node in every graph.  

We compare the test accuracy of our algorithm, OTC (short for Optimal Transport based Compression), to several state-of-the-art methods: REC \cite{LV2018}, Heavy edge matching (Heavy) \cite{DGK2007}, Affinity vertex proximity (Affinity) \cite{LB2012}, and Algebraic distance (Algebraic) \cite{RSB2011, CS2011}. 
Amongst these, the REC algorithm is a randomized method that iteratively contracts edges of the graph and thereby coarsens the graph. 
Since the method is randomized, REC yields a compressed graph that achieves a specified compression factor, i.e., the ratio of the number of nodes in the reduced graph to that in the original uncompressed graph, only in expectation. Therefore, in order to allow for a fair comparison, we  first run REC on each graph with the compression factor set  to 0.5, and then execute other baselines and our algorithm, i.e. Algorithm \ref{MPPrimal1}, with $k$ set to the number of nodes in the compressed graph produced by REC. Also to mitigate the effects of randomness on the number of nodes returned by REC, we performed 5 independent runs for REC, and subsequently all other methods.   

Each collection of compressed graphs was then divided into train and test sets, and used for our classification task. Since our datasets do not provide separate train and test sets, we employed the following procedure for each dataset. We partitioned each dataset into multiple train and test sets of varying sizes. Specifically, for each $p \in \{0.2, 0.3, 0.4, 0.5, 0.6, 0.7, 0.8\}$, we divided each dataset randomly into a train set containing a fraction $p$ of the graphs in the dataset, and a test set containing the remaining $1-p$ fraction. To mitigate the effects of chance, we formed 5 such independent train-test partitions for each fraction $p$ for each dataset. We averaged the results over these multiple splits to get one reading per collection, and thus 5 readings in total for all collections, for each fraction for each method. We averaged the test accuracy across collections for each method and fraction. 

We now specify the cost function $c$ for our algorithm. As described in section \ref{SecGraphCompress}, we can leverage the cost function to encode preference for different edges in the compressed graph. For each graph, we set $c=0.01$ for each edge $e$ incident on the nodes with same label, and $c=0.02$ for each $e$ incident on the nodes that have different label. Thus, in order to encourage diversity of labels, we slightly biased the graphs to prefer retaining the edges that have separate labels at their end-points. More generally, one could parameterize and thus learn the costs for edges. 

In our experiments, we employed support vector machines (SVMs), with Weisfeiler-Lehman subtree kernel \cite{SSLMB2011} to quantify the similarity between graphs \cite{NGBK2016, VSKB2010,  KP2016}. This kernel is based on the Weisfeiler-Lehman test of isomorphism \cite{WL1968, XHLJ2019}, and thus naturally takes into account the labels of the nodes in the two graphs. 
We fixed the number of kernel iterations to 5. We also fixed $T = 25$ for our algorithm. For each method and each train-test split, we used a separate 5-fold cross-validation procedure to tune the coefficient of error term $C$ over the set $\{0.1, 1, 10\}$ for training an independent SVM model on the training portion.  
\color{black}
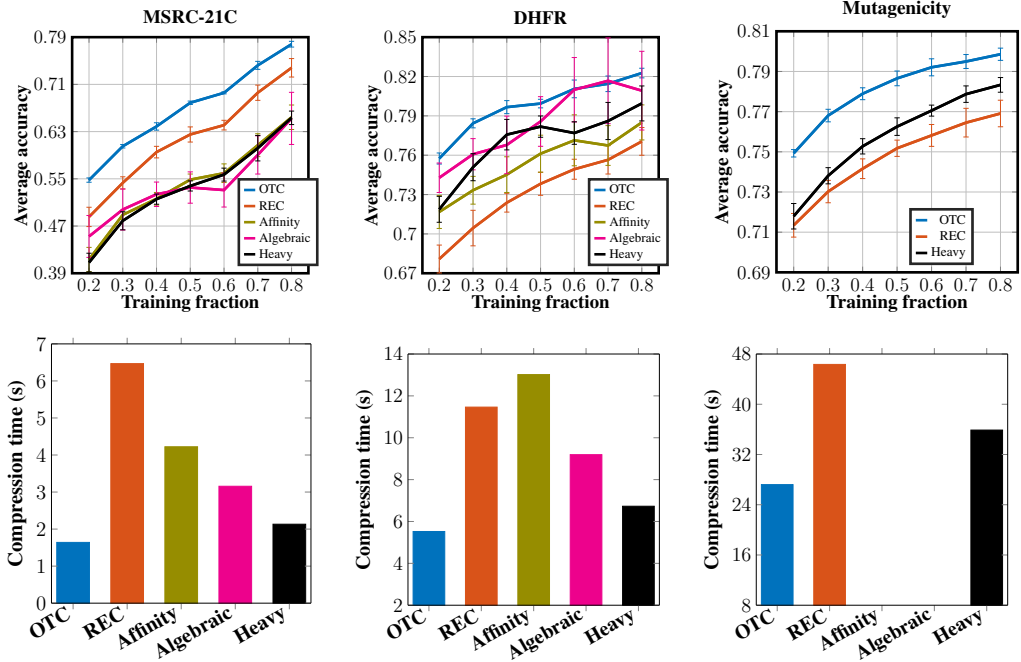
\begin{figure*}[t]
    \begin{subfigure}{}       
           \resizebox{.3\textwidth}{!}{\definecolor{mycolor1}{rgb}{0.00000,0.44700,0.74100}%
 \definecolor{mycolor2}{rgb}{0.85000,0.32500,0.09800} 
 \begin{tikzpicture} 
 \begin{axis}[%
 width= \figW, 
 height= \figH, 
 at={(1.011in,0.642in)}, 
 scale only axis, 
 xmin=0.15,  
 xmax=0.85,
 ymin=0.39,
 ymax=0.79,
 xlabel={{\bf Training fraction}},    
 ytick = {.39, .47, .55, .63, .71, .79, 0.86}, 
 yticklabel style={/pgf/number format/precision=3}, 
 line width=2pt,  
 grid=both,   
 grid style={line width=.3pt, draw=gray!10},  
 major grid style={line width=.2pt,draw=gray!50}, 
 ylabel={{\bf Average accuracy}},   
 axis background/.style={fill=white}, 
 title style= {font=\Large}, 
 title={{\bf MSRC-21C}}, 
 legend style={at={(0.7,0.03)},anchor= south west,legend cell align=left,align=left,draw=white!15!black}, 
 xlabel style={font=\Large},ylabel style={font=\Large}, ticklabel style={font=\Large},y label style={at={(axis description cs:-0.15,0.5)}},x label style={at={(axis description cs:.5,-0.07)}},legend style={legend cell align=left,align=left,draw=white!15!black},scaled y ticks = false, y tick label style={/pgf/number format/fixed}, legend image post style={scale=.5} 
 ] 
 
\addlegendentry{OTC}; 
\addplot [color=mycolor1,solid] table[row sep=crcr]{
0.8 0.778095238095\\
0.7 0.742222222222\\
0.6 0.695714285714\\
0.5 0.678857142857\\
0.4 0.63873015873\\
0.3 0.605170068027\\
0.2 0.547857142857\\
};

\addplot [color=mycolor1,solid,forget plot] 
                     plot [error bars/.cd, y dir = both, y explicit] 
                     table[row sep=crcr, y error plus index=2, y error minus index=3]{
0.8 0.778095238095 0.00485620906056 0.00485620906056\\
0.7 0.742222222222 0.00677909730288 0.00677909730288\\
0.6 0.695714285714 0.00178174161275 0.00178174161275\\
0.5 0.678857142857 0.0028507865804 0.0028507865804\\
0.4 0.63873015873 0.00623710562044 0.00623710562044\\
0.3 0.605170068027 0.00290532741552 0.00290532741552\\
0.2 0.547857142857 0.00364215679542 0.00364215679542\\
};

\addlegendentry{REC}; 
\addplot [color=mycolor2,solid] table[row sep=crcr]{
0.8 0.738095238095\\
0.7 0.695873015873\\
0.6 0.640952380952\\
0.5 0.625142857143\\
0.4 0.595238095238\\
0.3 0.543129251701\\
0.2 0.485476190476\\
};

\addplot [color=mycolor2,solid,forget plot] 
                     plot [error bars/.cd, y dir = both, y explicit] 
                     table[row sep=crcr, y error plus index=2, y error minus index=3]{
0.8 0.738095238095 0.0156492159287 0.0156492159287\\
0.7 0.695873015873 0.0129187237774 0.0129187237774\\
0.6 0.640952380952 0.00816496580928 0.00816496580928\\
0.5 0.625142857143 0.0128285396118 0.0128285396118\\
0.4 0.595238095238 0.0097847841317 0.0097847841317\\
0.3 0.543129251701 0.0103042700565 0.0103042700565\\
0.2 0.485476190476 0.0163472798988 0.0163472798988\\
};

\addlegendentry{Affinity}; 
\addplot [color=olive,solid] table[row sep=crcr]{
0.8 0.654285714286\\
0.7 0.606984126984\\
0.6 0.559523809524\\
0.5 0.548571428571\\
0.4 0.51619047619\\
0.3 0.488979591837\\
0.2 0.413333333333\\
};

\addplot [color=olive,solid,forget plot] 
                     plot [error bars/.cd, y dir = both, y explicit] 
                     table[row sep=crcr, y error plus index=2, y error minus index=3]{
0.8 0.654285714286 0.020778499266 0.020778499266\\
0.7 0.606984126984 0.0194040721142 0.0194040721142\\
0.6 0.559523809524 0.0157934513826 0.0157934513826\\
0.5 0.548571428571 0.00956182887468 0.00956182887468\\
0.4 0.51619047619 0.0106904496765 0.0106904496765\\
0.3 0.488979591837 0.00782046686433 0.00782046686433\\
0.2 0.413333333333 0.0205010922777 0.0205010922777\\
};

\addlegendentry{Algebraic}; 
\addplot [color=magenta,solid] table[row sep=crcr]{
0.8 0.652380952381\\
0.7 0.590476190476\\
0.6 0.530952380952\\
0.5 0.535238095238\\
0.4 0.52380952381\\
0.3 0.497959183673\\
0.2 0.452380952381\\
};

\addplot [color=magenta,solid,forget plot] 
                     plot [error bars/.cd, y dir = both, y explicit] 
                     table[row sep=crcr, y error plus index=2, y error minus index=3]{
0.8 0.652380952381 0.0441600880738 0.0441600880738\\
0.7 0.590476190476 0.0323747270704 0.0323747270704\\
0.6 0.530952380952 0.02876915708 0.02876915708\\
0.5 0.535238095238 0.0265302633851 0.0265302633851\\
0.4 0.52380952381 0.0206958806514 0.0206958806514\\
0.3 0.497959183673 0.0347937737484 0.0347937737484\\
0.2 0.452380952381 0.0355153187861 0.0355153187861\\
};

\addlegendentry{Heavy}; 
\addplot [color=black,solid] table[row sep=crcr]{
0.8 0.653333333333\\
0.7 0.601904761905\\
0.6 0.557142857143\\
0.5 0.537904761905\\
0.4 0.515555555556\\
0.3 0.478911564626\\
0.2 0.408095238095\\
};

\addplot [color=black,solid,forget plot] 
                     plot [error bars/.cd, y dir = both, y explicit] 
                     table[row sep=crcr, y error plus index=2, y error minus index=3]{
0.8 0.653333333333 0.0114285714286 0.0114285714286\\
0.7 0.601904761905 0.0215873015873 0.0215873015873\\
0.6 0.557142857143 0.0109627280316 0.0109627280316\\
0.5 0.537904761905 0.00856931190796 0.00856931190796\\
0.4 0.515555555556 0.00860077281533 0.00860077281533\\
0.3 0.478911564626 0.0152962931402 0.0152962931402\\
0.2 0.408095238095 0.0158364658555 0.0158364658555\\
};

\end{axis}
\end{tikzpicture}}
            \label{fig:Algorithms}
    \end{subfigure}%
    ~~
    \begin{subfigure}{}       
           \resizebox{.3\textwidth}{!}{\definecolor{mycolor1}{rgb}{0.00000,0.44700,0.74100}%
 \definecolor{mycolor2}{rgb}{0.85000,0.32500,0.09800} 
 \begin{tikzpicture} 
 \begin{axis}[%
 width= \figW, 
 height= \figH, 
 at={(1.011in,0.642in)}, 
 scale only axis, 
 xmin=0.15,  
 xmax=0.85,  
 ymin=0.67,
 ymax=0.85,
 xlabel={{\bf Training fraction}},    
 ytick = {.67, .70, .73, .76, .79, .82, 0.85}, 
 yticklabel style={/pgf/number format/precision=3}, 
 line width=2pt,  
 grid=both,   
 grid style={line width=.3pt, draw=gray!10},  
 major grid style={line width=.2pt,draw=gray!50}, 
 ylabel={{\bf Average accuracy}},   
 axis background/.style={fill=white}, 
 title style= {font=\Large}, 
 title={{\bf DHFR}}, 
 legend style={at={(0.7,0.03)},anchor= south west,legend cell align=left,align=left,draw=white!15!black}, 
 xlabel style={font=\Large},ylabel style={font=\Large}, ticklabel style={font=\Large},y label style={at={(axis description cs:-0.15,0.5)}},x label style={at={(axis description cs:.5,-0.07)}},legend style={legend cell align=left,align=left,draw=white!15!black},scaled y ticks = false, y tick label style={/pgf/number format/fixed}, legend image post style={scale=.5} 
 ] 
 


\addlegendentry{OTC}; 
\addplot [mycolor1,solid] table[row sep=crcr]{
0.8 0.822631578947\\
0.7 0.814449339207\\
0.6 0.810561056106\\
0.5 0.799365079365\\
0.4 0.796651982379\\
0.3 0.784301886792\\
0.2 0.757487603306\\
};

\addplot [mycolor1,solid,forget plot] 
                     plot [error bars/.cd, y dir = both, y explicit] 
                     table[row sep=crcr, y error plus index=2, y error minus index=3]{
0.8 0.822631578947 0.0036653653361 0.0036653653361\\
0.7 0.814449339207 0.005939136404 0.005939136404\\
0.6 0.810561056106 0.00675721921693 0.00675721921693\\
0.5 0.799365079365 0.00326845401301 0.00326845401301\\
0.4 0.796651982379 0.00497309807636 0.00497309807636\\
0.3 0.784301886792 0.00344699992352 0.00344699992352\\
0.2 0.757487603306 0.00421588061274 0.00421588061274\\
};

\addlegendentry{REC}; 
\addplot [color=mycolor2,solid] table[row sep=crcr]{
0.8 0.770526315789\\
0.7 0.756475770925\\
0.6 0.749306930693\\
0.5 0.738201058201\\
0.4 0.723788546256\\
0.3 0.704452830189\\
0.2 0.680859504132\\
};

\addplot [color=mycolor2,solid,forget plot] 
                     plot [error bars/.cd, y dir = both, y explicit] 
                     table[row sep=crcr, y error plus index=2, y error minus index=3]{
0.8 0.770526315789 0.0105788164433 0.0105788164433\\
0.7 0.756475770925 0.010813699378 0.010813699378\\
0.6 0.749306930693 0.00762254787234 0.00762254787234\\
0.5 0.738201058201 0.00877988085193 0.00877988085193\\
0.4 0.723788546256 0.00716316356799 0.00716316356799\\
0.3 0.704452830189 0.0135437531824 0.0135437531824\\
0.2 0.680859504132 0.0106312733843 0.0106312733843\\
};

\addlegendentry{Affinity}; 
\addplot [color=olive,solid] table[row sep=crcr]{
0.8 0.785\\
0.7 0.767400881057\\
0.6 0.771353135314\\
0.5 0.761164021164\\
0.4 0.745198237885\\
0.3 0.733283018868\\
0.2 0.716760330579\\
};

\addplot [color=olive,solid,forget plot] 
                     plot [error bars/.cd, y dir = both, y explicit] 
                     table[row sep=crcr, y error plus index=2, y error minus index=3]{
0.8 0.785 0.013356397411 0.013356397411\\
0.7 0.767400881057 0.0151582824089 0.0151582824089\\
0.6 0.771353135314 0.0194135725097 0.0194135725097\\
0.5 0.761164021164 0.014088785723 0.014088785723\\
0.4 0.745198237885 0.0136111229377 0.0136111229377\\
0.3 0.733283018868 0.0105859650363 0.0105859650363\\
0.2 0.716760330579 0.0126008966672 0.0126008966672\\
};

\addlegendentry{Algebraic}; 
\addplot [color=magenta,solid] table[row sep=crcr]{
0.8 0.809210526316\\
0.7 0.816740088106\\
0.6 0.809900990099\\
0.5 0.785714285714\\
0.4 0.767841409692\\
0.3 0.760754716981\\
0.2 0.742809917355\\
};

\addplot [color=magenta,solid,forget plot] 
                     plot [error bars/.cd, y dir = both, y explicit] 
                     table[row sep=crcr, y error plus index=2, y error minus index=3]{
0.8 0.809210526316 0.03000461645 0.03000461645\\
0.7 0.816740088106 0.0326823559313 0.0326823559313\\
0.6 0.809900990099 0.0245736111802 0.0245736111802\\
0.5 0.785714285714 0.018929676 0.018929676\\
0.4 0.767841409692 0.021964671123 0.021964671123\\
0.3 0.760754716981 0.0118972700154 0.0118972700154\\
0.2 0.742809917355 0.0112396694215 0.0112396694215\\
};

\addlegendentry{Heavy}; 
\addplot [color=black,solid] table[row sep=crcr]{
0.8 0.799473684211\\
0.7 0.786079295154\\
0.6 0.776897689769\\
0.5 0.781798941799\\
0.4 0.775682819383\\
0.3 0.750867924528\\
0.2 0.718743801653\\
};

\addplot [color=black,solid,forget plot] 
                     plot [error bars/.cd, y dir = both, y explicit] 
                     table[row sep=crcr, y error plus index=2, y error minus index=3]{
0.8 0.799473684211 0.0133304475178 0.0133304475178\\
0.7 0.786079295154 0.0141562627442 0.0141562627442\\
0.6 0.776897689769 0.00858593401532 0.00858593401532\\
0.5 0.781798941799 0.00812544077056 0.00812544077056\\
0.4 0.775682819383 0.01162661813 0.01162661813\\
0.3 0.750867924528 0.0103217032436 0.0103217032436\\
0.2 0.718743801653 0.0098807036226 0.0098807036226\\
};

\end{axis}
\end{tikzpicture}}
            \label{fig:Algorithms}
    \end{subfigure}%
    ~~ 
    \begin{subfigure}{}      
          \resizebox{.3\textwidth}{!}{\definecolor{mycolor1}{rgb}{0.00000,0.44700,0.74100}%
 \definecolor{mycolor2}{rgb}{0.85000,0.32500,0.09800} 
 \begin{tikzpicture} 
 \begin{axis}[%
 width= \figW, 
 height= \figH, 
 at={(1.011in,0.642in)}, 
 scale only axis, 
 xmin=0.15,  
 xmax=0.85,  
 ymin=0.69,
 ymax=0.81,
 xlabel={{\bf Training fraction}},    
 ytick = {0.69, .71, .73, .75, .77, .79, .81}, 
 yticklabel style={/pgf/number format/precision=3}, 
 line width=2pt,  
 grid=both,   
 grid style={line width=.3pt, draw=gray!10},  
 major grid style={line width=.2pt,draw=gray!50},
 ylabel={{\bf Average accuracy}},   
 axis background/.style={fill=white}, 
 title style= {font=\Large}, 
 title={{\bf Mutagenicity}}, 
 legend style={at={(0.55,0.03)},anchor= south west,legend cell align=left,align=left,draw=white!15!black}, 
 xlabel style={font=\Large},ylabel style={font=\Large}, ticklabel style={font=\Large},y label style={at={(axis description cs:-0.16,0.5)}},x label style={at={(axis description cs:.5,-0.07)}},legend style={legend cell align=right,draw=white!15!black},scaled y ticks = false, y tick label style={/pgf/number format/fixed}, legend image post style={scale=.5} 
 ] 
 
\addlegendentry{OTC}; 
\addplot [color=mycolor1,solid] table[row sep=crcr]{
0.8  0.7985714285714286\\
0.7 0.7949615975422428\\
0.6 0.7920691642651297\\
0.5 0.7865560165975103\\
0.4 0.7789166346523244\\
0.3 0.7680500658761529\\
0.2 0.7492680115273774\\
};

\addplot [color=mycolor1,solid,forget plot] 
                     plot [error bars/.cd, y dir = both, y explicit] 
                     table[row sep=crcr, y error plus index=2, y error minus index=3]{
0.8 0.7985714285714286  0.003044266279246792 0.003044266279246792 \\
0.7 0.7949615975422428 0.003469542513520508 0.003469542513520508\\
0.6 0.7920691642651297 0.004239442862182531 0.004239442862182531\\
0.5 0.7865560165975103 0.003665396624820227 0.003665396624820227\\
0.4 0.7789166346523244 0.002960907931755428 0.002960907931755428\\
0.3 0.7680500658761529 0.003122279471641328 0.003122279471641328\\
0.2 0.7492680115273774 0.0018452447542376175 0.0018452447542376175\\
};

\addlegendentry{REC}; 
\addplot [color=mycolor2,solid] table[row sep=crcr]{
0.8 0.76907834\\
0.7 0.76451613\\
0.6  0.75815562\\
0.5 0.75177501\\
0.4 0.74159047\\
0.3 0.7302108\\
0.2 0.71342939\\
};

\addplot [color=mycolor2,solid,forget plot] 
                     plot [error bars/.cd, y dir = both, y explicit] 
                     table[row sep=crcr, y error plus index=2, y error minus index=3]{
0.8 0.76907834  0.00660837 0.00660837\\
0.7 0.76451613 0.00712856 0.00712856\\
0.6 0.75815562  0.00544258 0.00544258\\
0.5 0.75177501 0.00405004 0.00405004\\
0.4 0.74159047 0.00490042 0.00490042\\
0.3 0.7302108  0.00556793 0.00556793\\
0.2 0.71342939 0.00588495 0.00588495\\
};

\addlegendentry{Heavy}; 
\addplot [color=black,solid] table[row sep=crcr]{
0.8 0.78331797\\
0.7 0.77870968\\
0.6 0.77051297\\
0.5 0.76254495\\
0.4 0.75273146\\
0.3 0.73808959\\
0.2 0.71792507\\
};

\addplot [color=black,solid,forget plot] 
                     plot [error bars/.cd, y dir = both, y explicit] 
                     table[row sep=crcr, y error plus index=2, y error minus index=3]{
0.8 0.78331797 0.00365075 0.00365075\\
0.7 0.77870968 0.00410641 0.00410641\\
0.6 0.77051297 0.00273799 0.00273799\\
0.5 0.76254495 0.00437632 0.00437632\\
0.4 0.75273146 0.00380013 0.00380013\\
0.3 0.73808959 0.00402856 0.00402856\\
0.2 0.71792507 0.00634226 0.00634226\\
};

\end{axis}
\end{tikzpicture}}
            \label{fig:Algorithms}
    \end{subfigure}%
    ~~
    \begin{subfigure}{}       
          \resizebox{.3\textwidth}{!}{\definecolor{mycolor1}{rgb}{0.00000,0.44700,0.74100}%
 \definecolor{mycolor2}{rgb}{0.85000,0.32500,0.09800}
 
    \begin{tikzpicture}
        \begin{axis}[
        ybar,
        ymin=0,
        ymax=7,
        clip=false,
        separate axis lines,
        axis on top,
        ytick = {0, 1, 2, 3, 4, 5, 6, 7}, 
        xlabel style={text width=5cm}, ylabel style={font=\Large},
        width= \figW, 
 height= \figH, 
 scale only axis, 
        bar width=0.8cm,
        ylabel={\bf Compression time (s)},
        ticklabel style={font=\Large},
        symbolic x coords={{\bf OTC-50}, {\bf OTC}, {\bf REC},{\bf Affinity},{\bf Algebraic},{\bf Heavy}},
        x tick label style={rotate=30, anchor=east, align=left},
        nodes near coords align={vertical}, 
        bar shift=0pt,
        ]
        \addplot[fill, color=mycolor1] coordinates {({\bf OTC},1.6406295776367188)};
        \addplot[fill, color=mycolor2] coordinates {({\bf REC},6.4674536228179935)};
        \addplot[fill, color=olive] coordinates {({\bf Affinity},4.22278242111206)};
        \addplot[fill, color=magenta] coordinates {({\bf Algebraic},3.1566306591033935)};
        \addplot[fill, color=black] coordinates {({\bf Heavy},2.132055234909058)};
        \end{axis}
        \end{tikzpicture} 
 

}
            \label{fig:Algorithms}
    \end{subfigure}%
    ~~
    \begin{subfigure}{}       
          \resizebox{.3\textwidth}{!}{\definecolor{mycolor1}{rgb}{0.00000,0.44700,0.74100}%
 \definecolor{mycolor2}{rgb}{0.85000,0.32500,0.09800}
 
    \begin{tikzpicture}
        \begin{axis}[
        ybar,
        ymin=2,
        ymax=14,
        clip=false,
        separate axis lines,
        axis on top,
        ytick = {2, 4, 6, 8, 10, 12, 14}, 
        xlabel style={text width=5cm}, ylabel style={font=\Large},
        width= \figW, 
 height= \figH, 
 scale only axis, 
        bar width=0.8cm,
        ylabel={\bf Compression time (s)},
        ticklabel style={font=\Large},
        symbolic x coords={{\bf OTC-50}, {\bf OTC}, {\bf REC},{\bf Affinity},{\bf Algebraic},{\bf Heavy}},
        x tick label style={rotate=30, anchor=east, align=left},
        nodes near coords align={vertical}, 
        bar shift=0pt,
        ]
        \addplot[fill, color=mycolor1] coordinates {({\bf OTC},5.522381782531738)};
        \addplot[fill, color=mycolor2] coordinates {({\bf REC},11.46413025856018)};
        \addplot[fill, color=olive] coordinates {({\bf Affinity},13.017185974121094)};
        \addplot[fill, color=magenta] coordinates {({\bf Algebraic},9.192070055007935)};
        \addplot[fill, color=black] coordinates {({\bf Heavy},6.729036140441894)};
        \end{axis}
        \end{tikzpicture} 
 

}
            \label{fig:Algorithms}
    \end{subfigure}%
    ~~
    \begin{subfigure}{}       
          \resizebox{.3\textwidth}{!}{\definecolor{mycolor1}{rgb}{0.00000,0.44700,0.74100}%
 \definecolor{mycolor2}{rgb}{0.85000,0.32500,0.09800}
 
    \begin{tikzpicture}
        \begin{axis}[
        ybar,
        ymin=8,
        ymax=48,
        clip=false,
        separate axis lines,
        axis on top,
        ytick = {8, 16, 24, 32, 40, 48}, 
        xlabel style={text width=5cm}, ylabel style={font=\Large},
        width= \figW, 
 height= \figH, 
 scale only axis, 
        bar width=0.8cm,
        ylabel={\bf Compression time (s)},
        ticklabel style={font=\Large},
        symbolic x coords={{\bf OTC-50}, {\bf OTC}, {\bf REC},{\bf Affinity},{\bf Algebraic},{\bf Heavy}},
        x tick label style={rotate=30, anchor=east, align=left},
        nodes near coords align={vertical}, 
        bar shift=0pt,
        ]
        \addplot[fill, color=mycolor1] coordinates {({\bf OTC},27.206649589538575)};
        \addplot[fill, color=mycolor2] coordinates {({\bf REC},46.34483013153076)};
        \addplot[fill, color=black] coordinates {({\bf Heavy},35.89169197082519)};
        \end{axis}
        \end{tikzpicture} 
 

}
            \label{fig:Algorithms}
    \end{subfigure}%
 \caption{{\bf Comparison on standard graph datasets}. The top row shows the average test accuracy and corresponding standard deviation for our method (OTC) and state-of-the-art baselines for different fractions of training data. 
 The bottom row compares the corresponding compression times. Our method outperforms the other methods in terms of both accuracy and compression time.  
 \label{fig:AlOu}}
\end{figure*}
Table \ref{tab:compare} summarizes the performance of different methods for each fraction of the data. Clearly, as the numbers in bold indicate, our method generally outperformed the other methods across the datasets. We note that on some datasets the discrepancy between the average test accuracy of two algorithms is massive for every fraction. 
Further, as Fig. \ref{fig:AlOu} shows, OTC performed best  in terms of compression time as well. We emphasize that the discrepancy in compression times became quite stark for larger datasets (i.e., DHFR and Mutagenicity). 
\begin{figure*}[t!]%
\hskip -0.5cm
\subfigure[Synthetic graph]{%
\label{fig:first}%
\includegraphics[width=1.9in, height=2in]{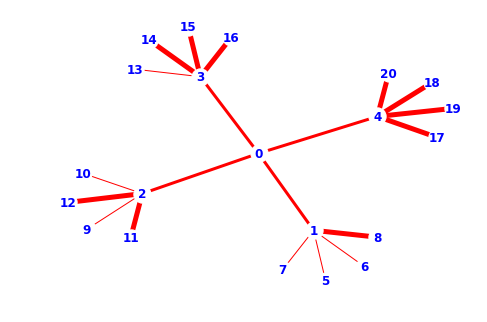}}%
~
\hskip -0.5cm
\subfigure[Compressed graph ($k=20$)]{%
\label{fig:second}%
\includegraphics[width=1.9in, height=2in]{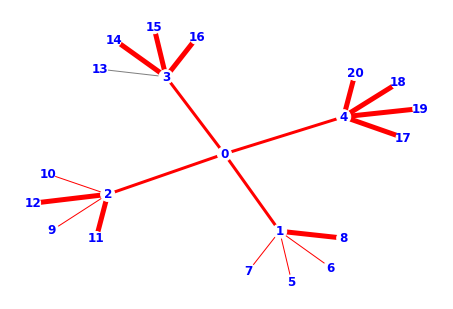}}%
~
\hskip -0.5cm
\subfigure[Compressed graph ($k=15$)]{%
\label{fig:second}%
\includegraphics[width=1.9in, height=2in]{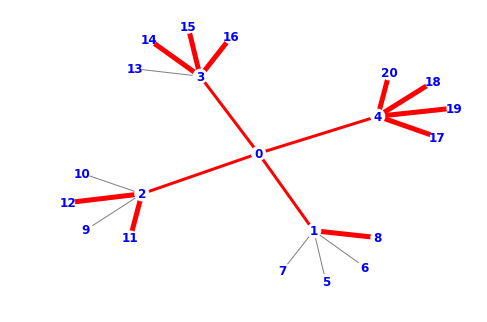}}%

\hskip -0.5cm
\subfigure[Compressed graph ($k=5$)]{%
\label{fig:third}%
\includegraphics[width=1.9in, height=2in]{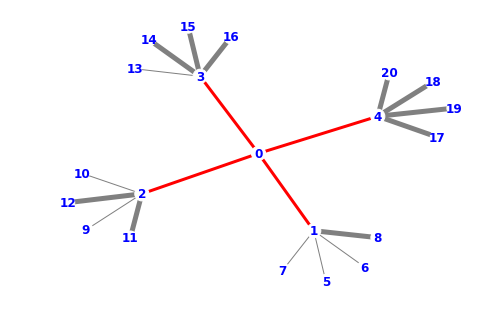}}%
~
\hskip -0.5cm
\subfigure[Mutagenicity]{%
\label{fig:fourth}%
\includegraphics[width=2in, height=2in]{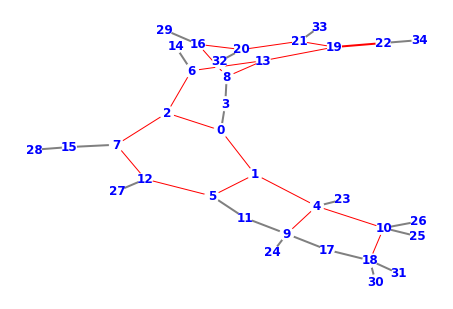}}%
~
\subfigure[MSRC-21C]{%
\label{fig:sixth}%
\includegraphics[width=1.8in, height=2in]{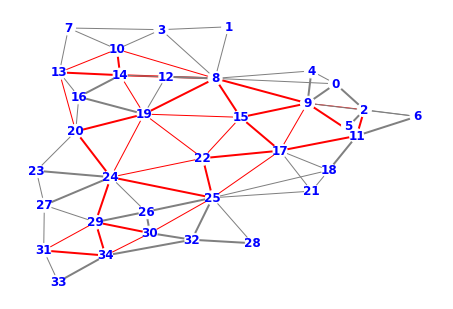}}
\caption{{\bf Visualizing compression on synthetic and real examples.} (a) A synthetic graph structured as a $4$-ary tree of depth $2$. The root $0$ is connected to its neighbors by edges having $c(e)=0.3$. All the other edges have either $c(e)=0.5$ (thickest) or $c(e)=0.1$ (lightest). The left out portions are shown in gray. (b) Leaf node 13, which is in the same subtree as three nodes with heavy edges 14-16, is the first to go.  (c) Proceeding further, 9 and 10 are left out followed by the remaining nodes (i.e. 5, 6,7) connected by light edges. (d) When the graph is compressed to 5 vertices, only the root and its neighbors remain despite bulkier subtrees, e.g. the one with node 4 and its neighbors, that are discarded. Thus, our method yields meaningful compression on this synthetic example. (e-f) Compressed structures pertaining to some sample graphs from real datasets that have some other interesting motifs. In each case, the compressed graph consists of red edges and the incident vertices, while the discarded parts are shown in gray. All the figures here are best viewed in color. 
\label{fig:fig_visual}}
\end{figure*}
\subsection{Compressing synthetic and real examples}
We now describe our second set of experiments with both synthetic and real data to show that our method can be seeded with useful prior information toward preserving interesting patterns in the compressed structures. This flexibility in specifying the prior information makes our approach especially well-suited for downstream tasks, where domain knowledge is often available.  

Fig. \ref{fig:fig_visual} demonstrates the effect of compressing a synthetic tree-structured graph. The penultimate level consists of four nodes, each of which has four child nodes as leaves. We introduce asymmetry with respect to the different nodes by specifying different combinations of $c(e)$ for edges $e$ between the leaf nodes and their parents: there are respectively one, two, three and four heavy edges (i.e. with $c(e) = 0.5$) from the internal nodes 1, 2, 3, and 4 to their children, i.e., the leaves in their subtree. As shown in the figure, our method adapts to the hierarchical structure with a change in the amount of compression from just one node to about three-fourths of the entire graph. We also show meaningful compression on some examples from real datasets. The bottom row of Fig. \ref{fig:fig_visual} shows two such examples, one each from Mutagenicity and MSRC-21C. For these graphs, we used the same specification for $c$ as in  section \ref{ClassifySection}. The example from Mutagenicity contains patterns such as rings and backbone structures that are ubiquitous in molecules and proteins. Likewise, the other example is a good representative of the MSRC-21C dataset from computer vision. Thus, our method encodes prior information, and provides fast and effective graph compression for downstream applications.  

\bibliography{main}
\bibliographystyle{unsrt}
\appendix
\clearpage
\section{Supplementary Material}
We provide here proofs of all the results from the main text. \\ \\
\textbf{Proof of Theorem \ref{Theorem1}}
\begin{proof}
We will prove the result by contradiction. 

Suppose there is some $\hat{e} \in E$ such that the optimal solution assigns $J^+(\hat{e}) > 0$ and $J^-(\hat{e}) > 0$. Then we note that
$$a ~\triangleq~ \min\{J^+(\hat{e}), J^-(\hat{e})\} > 0~.$$ 
Consider an alternate solution $(\tilde{J}^+, \tilde{J}^-)$ such that $\forall~ e  \in E$,
$$
\tilde{J}^+(e) ~=~
\begin{cases}
J^+(e) - a & \mbox{ if }  e = \hat{e}\\
J^+(e) & \mbox{ if }  e \in E \setminus \{\hat{e}\}\\
\end{cases}~,
$$
$$
\tilde{J}^-(e) ~=~
\begin{cases}
J^-(e) - a & \mbox{ if }  e = \hat{e}\\
J^-(e) & \mbox{ if }  e \in E \setminus \{\hat{e}\}\\
\end{cases}~.
$$
\noindent Clearly, $(\tilde{J}^+, \tilde{J}^-)$ is feasible for \eqref{UnWass}. Moreover, it achieves a lower value of the objective than the  optimal solution $(J^+, J^-)$. Therefore,  $(J^+, J^-)$ cannot be optimal.   
\end{proof}

\textbf{Proof of Theorem \ref{Theorem2}}
\begin{proof}
We introduce non-negative Lagrangian vectors $\alpha$ and $\beta$, respectively, for the constraints $Ft \preceq c$ and $-c \preceq Ft$. We consider the terms in the objective that depend on $t$ 
$$g(t) = t^{\top}(\rho_1 - \rho_0)  + \alpha^{\top} (Ft-c) - \beta^{\top} (Ft+c)~.$$ 
The gradient $\nabla g(t)$ must vanish at optimality, so
$$\rho_1^* - \rho_0 + F^{\top}(\alpha - \beta)~=~ \boldsymbol{0}~.$$
The first part of the theorem follows immediately by defining $\eta = \beta - \alpha$. A closer look at \eqref{UnWass} reveals that  $\eta = J^{-} - J^+$ is, in fact, the net flow along the edges $e^-$ from \eqref{UnWass}.  

Now, we prove the second part. By definition, in order for an edge $e$ to be active, at least one of $e^+$ and $e^-$ must be active, i.e., we must have $J^+(e) + J^-(e) > 0$. On the other hand, Theorem \ref{Theorem1} implies that at least one of $J^+(e)$ and $J^-(e)$ is 0 for each $e \in E$ in the optimal solution. Combining these facts, we have that for any active edge $e$, exactly one of   $e^+$ and $e^-$ is active, i.e.,  exactly one of the inequalities $J^+(e) > 0$ and $J^-(e) > 0$ must hold. This immediately implies, by complementary slackness, that exactly one of $\alpha(e)$ or $\beta(e)$ is 0. Thus, for any active edge $e$, either the lower bound or the upper bound on $Ft^*(e)$ in the constraints $- c(e) \leq Ft^*(e)  \leq  c(e)$ must become tight. Therefore, we must have $Ft^*(e) \in \{\pm c(e)\}$. 
 \end{proof}

\textbf{Proof of Theorem \ref{Theorem6}}
\begin{proof}
Note that since at least one coordinate of $\epsilon$ is strictly greater than 0, the feasible region is non-empty, and consequently, a unique projection exists. We introduce variables $\alpha \in \mathbb{R}$ and $\beta \in \mathbb{R}_+^d$, and form the Lagrangian
$$L(x, \alpha, \beta) ~=~   \dfrac{1}{2} ||x - y||^2 - \alpha ((x \odot \epsilon)^{\top} \boldsymbol{1} ~-~1) ~-~ \beta^{\top} (x \odot \epsilon)~.$$
We now write the KKT conditions for the optimal solution $x$. For each $j \in [d]$, we must have
\begin{eqnarray*}
x_j - y_j - \alpha \epsilon_j - \beta_j \epsilon_j & = & ~~0 \\
\epsilon_j x_j & \geq & ~~0 \\
\beta_j & \geq & ~~0 \\
\epsilon_j x_j \beta_j & = & ~~0~.\\
\text {Additionally}, \qquad \sum_{j=1}^d \epsilon_j x_j & = & ~~1~. 
\end{eqnarray*}
Clearly, for $j \in [d] \triangleq \{1, 2, \ldots, d\}, \epsilon_j = 0 \implies x_j = y_j$. Therefore, without loss of generality we assume in the rest of the proof that $\epsilon_j > 0$ for all $j$.  Then we can immediately simplify the KKT conditions to 
\begin{eqnarray}
x_j - y_j - \alpha \epsilon_j - \beta_j \epsilon_j & = & ~~0 \label{KKT1} \\
x_j & \geq & ~~0 \label{KKT2} \\
\beta_j & \geq & ~~0~~~.  \label{KKT3}\\
x_j \beta_j & = & ~~0~ \label{KKT4} \\
\sum_{j=1}^d \epsilon_j x_j & = & ~~1~ \label{KKT5}
\end{eqnarray}
We note that $x_j > 0 \xRightarrow[{}]{\eqref{KKT4}} \beta_j = 0 \xRightarrow[{}]{\eqref{KKT1}} y_j + \alpha \epsilon_j > 0 \xRightarrow[{}]{\epsilon_j > 0} y_j/\epsilon_j > -\alpha~,$
whereas 
\begin{eqnarray*}
x_j = 0 \xRightarrow[{}]{\eqref{KKT1}} y_j + \alpha \epsilon_j = -\beta_j \epsilon_j  \xRightarrow[{\epsilon_j > 0}]{\eqref{KKT3}} y_j + \alpha \epsilon_j \leq 0  \xRightarrow[{}]{\epsilon_j > 0} y_j/\epsilon_j \leq -\alpha ~.
\end{eqnarray*}
This shows that the zero coordinates $x_j$ correspond to smaller values of $y_j/\epsilon_j$. Thus, we can sort the indices $j$ in non-increasing order based on the ratio $y_j/\epsilon_j$, reorder $x$ according to the sorted indices, and find an index $\ell \in [d]$ such that 
$x_j > 0$ for $j \in [\ell]$ and 0 for $\ell < j \leq d$.  Without loss of generality, we therefore assume that 
\begin{eqnarray} x_1 \geq x_2 \ldots \geq x_{\ell} > 0 = x_{\ell + 1} \ldots = x_{d}~, \text{ and } \nonumber \\  y_1/\epsilon_1 \geq y_2/\epsilon_2 \ldots \geq y_d/\epsilon_d~.  \label{sortedorder} \end{eqnarray}
We then have from \eqref{KKT5} that 
\begin{eqnarray}  1 ~=~ \sum_{j=1}^d \epsilon_j x_j ~=~ \sum_{j=1}^{\ell} \epsilon_j x_j ~=~  \sum_{j=1}^{\ell}  \epsilon_j (y_j + \alpha \epsilon_j) \nonumber \\ \implies \alpha = \dfrac{1 - \sum_{j=1}^{\ell}  \epsilon_j y_j}{ \sum_{j=1}^{\ell} \epsilon_j^2}~~. \label{Lang_alpha} 
\end{eqnarray}
Thus, our task essentially boils down to finding the number of positive coordinates $\ell$.  We now show that 
$$\ell = \max\left\{j \in [d] ~ \bigg| ~ y_j + \epsilon_j \dfrac{(1 - \sum_{i=1}^j \epsilon_i y_i)}{\sum_{i=1}^j \epsilon_i^2} > 0\right\}~.$$
First consider $j < \ell$. Then $y_j/\epsilon_{j} > -\alpha$ for $j \in [\ell]$. Noting that $\epsilon_j > 0$ for all $j$ and using \eqref{Lang_alpha}, we must have
\begin{eqnarray*}
&&  y_j + \epsilon_j \dfrac{(1 - \sum_{i=1}^j \epsilon_i  y_i)}{\sum_{i=1}^j \epsilon_i^2} \\ & = & \dfrac{\epsilon_j}{\sum_{i=1}^j \epsilon_i^2} \left(y_j \dfrac{\sum_{i=1}^j \epsilon_i^2}{\epsilon_j}  + 1 - \sum_{i=1}^j \epsilon_i y_i \right)\\
&& \\
&&\noindent \text{which has the same sign as}  \\
&  & y_j \dfrac{\sum_{i=1}^j \epsilon_i^2}{\epsilon_j}  + 1 - \sum_{i=1}^j \epsilon_i y_i \\
& = & y_j \dfrac{\sum_{i=1}^j \epsilon_i^2}{\epsilon_j}  + \sum_{i=j+1}^{\ell} \epsilon_i y_i  + 1 - \sum_{i=1}^{\ell} \epsilon_i y_i \\
& = & y_j \dfrac{\sum_{i=1}^j \epsilon_i^2}{\epsilon_j}  + \sum_{i=j+1}^{\ell} \epsilon_i y_i  + \alpha \sum_{i=1}^{\ell} \epsilon_i^2 ~\qquad  \\
& = &  \left(\dfrac{y_j}{\epsilon_j} + \alpha\right) \sum_{i=1}^j \epsilon_i^2  + \sum_{i=j+1}^{\ell} \epsilon_i ^2\left(\dfrac{y_i}{\epsilon_i}  + \alpha \right) \\
& > & 0~. 
\end{eqnarray*}
Now consider $j = \ell$. Since $y_{\ell}/\epsilon_{\ell} > -\alpha$ and $\epsilon_{\ell} > 0$, we have $y_\ell + \alpha \epsilon_\ell > 0$. Thus
\begin{eqnarray*}
y_j + \epsilon_j \dfrac{(1 - \sum_{i=1}^j \epsilon_i y_i)}{\sum_{i=1}^j \epsilon_i^2}  & = & y_{\ell} + \epsilon_{\ell} \dfrac{(1 - \sum_{i=1}^{\ell} \epsilon_i y_i)}{\sum_{i=1}^{\ell} \epsilon_i^2} \\ & = &y_{\ell} +  \alpha \epsilon_{\ell} ~>~ 0~.
\end{eqnarray*}
Finally, we consider $\ell < j \leq d$. We note that
\begin{eqnarray*}
&& y_j + \epsilon_j \dfrac{(1 - \sum_{i=1}^j \epsilon_i y_i)}{\sum_{i=1}^j \epsilon_i^2}\\ & = & \dfrac{\epsilon_j}{\sum_{i=1}^j \epsilon_i^2} \left(y_j \dfrac{\sum_{i=1}^j \epsilon_i^2}{\epsilon_j}  + 1 - \sum_{i=1}^j \epsilon_i y_i \right)~,\\
&& \\
&&\noindent \text{which has the same sign as}  \\
&& y_j \dfrac{\sum_{i=1}^j \epsilon_i^2}{\epsilon_j}  + 1 - \sum_{i=1}^j \epsilon_i y_i \\
& = & y_j \dfrac{\sum_{i=1}^j \epsilon_i^2}{\epsilon_j}  + 1 - \sum_{i=1}^\ell \epsilon_i y_i - \sum_{i=\ell + 1}^j \epsilon_i y_i \\
& = & y_j \dfrac{\sum_{i=1}^j \epsilon_i^2}{\epsilon_j}  + \alpha \sum_{i=1}^\ell \epsilon_i^2 - \sum_{i=\ell + 1}^j \epsilon_i y_i ~\qquad \\
& = & \left(\dfrac{y_j}{\epsilon_j} + \alpha\right) \sum_{i=1}^\ell \epsilon_i^2  + \sum_{i=\ell+1}^{j}  \epsilon_i ^2 \left(\dfrac{y_j}{\epsilon_j} - \dfrac{y_i}{\epsilon_i} \right)~,\\
& \leq & 0~,
\end{eqnarray*}
by leveraging the sorted property in \eqref{sortedorder} and the fact that $y_j/\epsilon_{j}  \leq -\alpha$ for $j \in [\ell]$.

\noindent Therefore, we have shown that $y_j + \epsilon_j \dfrac{(1 - \sum_{i=1}^j \epsilon_i y_i)}{\sum_{i=1}^j \epsilon_i^2} > 0$ for all $j \in [\ell]$, and at most 0 for $\ell < j \leq d$. Algorithm  \ref{ProjectWS} implements this procedure, and that proves its correctness. The $O(d \log d)$ time complexity is due to the cost of sorting the indices $j \in [d]$ based on $y_j/\epsilon_j$.     
\end{proof}

\textbf{Proof of Theorem \ref{RelaxationExactness}}
\begin{proof}
Recall the formulation \eqref{EqP1_past}:
$$\min_{\substack{\epsilon \in \mathcal{E}_{k} \\ \tilde{\rho_1} \odot \epsilon \in \Delta(V)}} \max_{t \in \mathcal{T}_{F, c}} \underbrace{t^{\top}(\tilde{\rho}_1 \odot \epsilon - \rho_0) ~+~ \dfrac{\lambda}{2} ||\tilde{\rho}_1||^2}_{\phi(\epsilon, t, \tilde{\rho}_1)}~.$$
Making the constraints $\mathcal{E}_k$ explicit,   we get
\begin{eqnarray} \label{ExactEq1} \min_{\substack{\epsilon \in \{0, 1\}^{|V|} \\ \epsilon^{\top}\boldsymbol{1} \leq k}}~ \left(\min_{\substack{\tilde{\rho}_1 \in \mathbb{R}^{|V|}\\ \tilde{\rho}_1 \odot \epsilon \in \Delta(V)}} ~ \max_{t \in \mathcal{T}_{F, c}}~~ \phi(\epsilon, t, \tilde{\rho}_1)   \right)~.
\end{eqnarray}
Note that for any fixed $\epsilon$ (a) $\{\tilde{\rho}_1 \in \mathbb{R}^{|V|}~|~ (\tilde{\rho_1} \odot \epsilon) \in \Delta(V) \}$ is convex, and  $\mathcal{T}_{F, c}$ is convex and compact, (b)  $\phi(\epsilon, t, \tilde{\rho}_1)$ is continuous, and (c) for every fixed $t$,  $\phi(\epsilon, t, \cdot)$ is convex in $\tilde{\rho}_1$; while for every fixed $\tilde{\rho}_1$, $\phi(\epsilon, \cdot, \tilde{\rho}_1)$ is linear (thus concave) in $t$. Therefore, invoking the Sion's minimax theorem \cite{S1958}, we can swap the order of $\min$ and $\max$ within the parentheses in \eqref{ExactEq1}, and obtain 
\begin{eqnarray} \label{EqChange}
\min_{\substack{\epsilon \in \{0, 1\}^{|V|} \\ \epsilon^{\top}\boldsymbol{1} \leq k}}~~~ \left(\max_{t \in \mathcal{T}_{F, c}} \min_{\substack{\tilde{\rho}_1 \in \mathbb{R}^{|V|}\\ \tilde{\rho}_1 \odot \epsilon \in \Delta(V)}} \phi(\epsilon, t, \tilde{\rho}_1)   \right)~.
\end{eqnarray}
We introduce Lagrangian variables $\nu \in \mathbb{R}^{|V|}_+$ and $\zeta \in \mathbb{R}$, respectively, for the simplex constraints (a) $\boldsymbol{0} \preceq \tilde{\rho}_1 \odot \epsilon $ and (b) $(\tilde{\rho}_1 \odot \epsilon)^{\top} \boldsymbol{1} = 1$ to get
\begin{equation}  \label{PhiWithRho}
\Phi(\epsilon, t, \tilde{\rho}_1, \nu, \zeta) \triangleq \phi(\epsilon, t, \tilde{\rho}_1) + \zeta((\tilde{\rho}_1 \odot \epsilon)^{\top} \boldsymbol{1} - 1) - \nu^{\top}(\tilde{\rho}_1 \odot \epsilon)~~.
\end{equation}
Applying the optimality conditions, we note for any fixed pair $(\epsilon, t)$, the corresponding optimal $\tilde{\rho}_1$  must satisfy
\begin{equation*} 
\partial_{\tilde{\rho}_1} ~ \Phi(\epsilon, t, \tilde{\rho}_1, \nu, \zeta)  ~=~ 0~,
\end{equation*}
whereby 
\begin{equation} \label{hatrho}
\tilde{\rho}_1 ~=~  -~\left(\dfrac{(t - \nu) \odot \epsilon  + \zeta \epsilon}{\lambda}\right)~. 
\end{equation}
Plugging in $\tilde{\rho}_1$ from \eqref{hatrho} into  \eqref{PhiWithRho}, we thus have the following equivalent dual formulation for \eqref{EqChange}
\begin{eqnarray*} 
\min_{\substack{\epsilon \in \{0, 1\}^{|V|} \\ \epsilon^{\top}\boldsymbol{1} \leq k}}~~\max_{t \in \mathcal{T}_{F, c}} ~~\max_{\nu \in \mathbb{R}^{|V|}_+, \zeta \in \mathbb{R}} ~~\mathcal{M}(\epsilon, t, \nu, \zeta) ~,
\end{eqnarray*} 
where 
\begin{equation} \label{LagDual}
\mathcal{M}(\epsilon, t, \nu, \zeta) =    \dfrac{-1}{2\lambda} ||(t - \nu + \zeta \boldsymbol{1})\odot \epsilon||^2 - t^{\top} \rho_0 - \zeta~. 
\end{equation}
Invoking the first order convex optimality condition for constrained optimization, the $\hat{\epsilon}$ obtained from relaxation of \eqref{EqP1_past} is optimal if and only if 
 \begin{equation} \label{PartialE}
 \boldsymbol{0}  ~\in~  \left\{\partial_{\epsilon} \underbrace{\max_{t \in \mathcal{T}_{F, c}} ~~\max_{\nu \in \mathbb{R}^{|V|}_+, \zeta \in \mathbb{R}} ~~\mathcal{M}(\epsilon, t, \nu, \zeta)}_{(A)} ~+~ \mathbb{N}\right\},
 \end{equation}
 where $\mathbb{N}$ is the normal cone of the relaxed constraints 
 $$\tilde{\mathcal{E}}_k = 
 \left\{\epsilon \in [0, 1]^{|V|} ~\bigg|~ \epsilon^{\top} \boldsymbol{1} \leq k\right\}~.$$ 
 
Now we note that for any vector $x$, we can write $x \odot \epsilon =  D(\epsilon) x$, where $D(\epsilon)$ is the diagonal matrix corresponding to $\epsilon$.  Also, since $\epsilon \in \{0, 1\}^{|V|}$, we get $D(\epsilon)^{\top} D(\epsilon) = D(\epsilon)$. 
Thus, for any $x$, we have

$$||x \odot \epsilon||^2 =  ||D(\epsilon) x||^2 = (D(\epsilon) x)^{\top} D(\epsilon) x = x^{\top} D(\epsilon)^{\top} D(\epsilon) x = x^{\top}D(\epsilon)x~.$$

In particular, we can simplify $||(t - \nu + \zeta \boldsymbol{1}) \odot \epsilon||^2$ in \eqref{LagDual} when we set $x$ to $t - \nu + \zeta \boldsymbol{1}$. The theorem statement then follows immediately from \eqref{PartialE} by representing $\mathbb{N}$ at the integral point $\epsilon^*$ and leveraging the non-negative dual parameter associated with  the constraint $\epsilon^{\top} \boldsymbol{1} \leq k$. 
\end{proof}

\textbf{Proof of Theorem \ref{DualForm}}
\begin{proof}
We will use the shorthand $\tilde{\rho}_{1v}$ for $\tilde{\rho}_1(v)$, and likewise for indexing $t$, $\nu$, and $\epsilon$. Using \eqref{hatrho}, 
\begin{eqnarray*}
\tilde{\rho}_1 & =   -~\left(\dfrac{(t - \nu) \odot \epsilon  + \zeta \epsilon}{\lambda}\right) & =  -~\left(\dfrac{t - \nu + \zeta \boldsymbol{1}}{\lambda}\right) \odot \epsilon~.
\end{eqnarray*}
Therefore, 
\begin{equation} 
\label{rhoforKKT}
\tilde{\rho}_1 \odot \epsilon  =  -~\left(\dfrac{t - \nu + \zeta \boldsymbol{1}}{\lambda}\right) \odot \epsilon \odot \epsilon  ~~=~` -~\left(\dfrac{t - \nu + \zeta \boldsymbol{1}}{\lambda}\right) \odot \epsilon~~~ =~~ \tilde{\rho}_1~, 
\end{equation}
since $\epsilon \in \{0, 1\}^{|V|}$.  We write one of the KKT conditions for optimality
\begin{eqnarray*}
 \tilde{\rho}_1 \odot \epsilon \odot \nu & = & \tilde{\rho}_1 \odot \nu ~~=~~ \boldsymbol{0}~.
\end{eqnarray*}
We consider the different cases. Note that for $v \in V$, using \eqref{rhoforKKT}, we have
\begin{equation} \label{rho1condition1}
 \tilde{\rho}_{1v}  ~>~ 0  \implies  \nu_v = 0 \implies \tilde{\rho}_{1v} ~=~ - \dfrac{(t_v + \zeta) \epsilon_v }{\lambda}~,
 \end{equation}
\begin{equation} \label{rho1condition2} \tilde{\rho}_{1v}  ~=~ 0  \implies 
(t_v - \nu_v + \zeta)\epsilon_v ~=~ 0 \implies \nu_v \epsilon_v ~=~  (t_v + \zeta) \epsilon_v~.\end{equation}
But since $\nu_v \geq 0$, and $\epsilon_v \in \{0, 1\}$, we note that $\nu_v \epsilon_v \geq 0$. Then, by \eqref{rho1condition2}, we have
\begin{equation} \label{rho1condition2a} \nu_v \epsilon_v \geq 0 \implies  - \dfrac{(t_v + \zeta) \epsilon_v }{\lambda} ~\leq~ 0 ~ =~ \tilde{\rho}_{1v}~. \end{equation}
Combining \eqref{rho1condition1} and \eqref{rho1condition2a}, we can write 
\begin{eqnarray} \label{rho1final}
\tilde{\rho}_{1v} &=&  \max\left\{-\dfrac{(t_v + \zeta) \epsilon_v}{\lambda}, ~0\right\}~ ~=~
\dfrac{\epsilon_v}{\lambda} \max\left\{-(t_v + \zeta), ~0\right\}~,
\end{eqnarray}
since $\epsilon_v \geq 0$ for all $v \in V$ and $\lambda > 0$.
Therefore, we get $\tilde{\rho}_1 = \dfrac{\epsilon}{\lambda} \odot r_+$, where 
$r ~=~ -~(t + \zeta \boldsymbol{1})$, and $r_+$ is computed by setting the negative coordinates of $r$ to 0.\\ 

Moreover, since $\tilde{\rho}_1 \odot \nu ~=~ \boldsymbol{0}$, we can eliminate $\nu$ from \eqref{PhiWithRho} and write \eqref{ExactEq1} as
\begin{eqnarray*}
\min_{\epsilon \in \mathcal{E}_{k}}~\max_{\substack{t \in \mathcal{T}_{F, c}}}\max_{\zeta \in \mathbb{R}}  ~~~~ t^{\top}(\tilde{\rho}_1 - \rho_0) + \dfrac{\lambda}{2} ||\tilde{\rho}_1||^2 + \zeta(\tilde{\rho}_1^{\top}\boldsymbol{1} - 1)
\end{eqnarray*}
Substituting for $\tilde{\rho}_1$ from \eqref{rho1final},  we obtain the following equivalent problem
\begin{eqnarray*}
& \displaystyle \min_{\epsilon \in \mathcal{E}_{k}}~\max_{\substack{t \in \mathcal{T}_{F, c}}}\max_{\zeta \in \mathbb{R}}  ~~~~  - \dfrac{r^{\top}}{\lambda}(\epsilon \odot r_+)
~+~ \dfrac{1}{2\lambda} r_+^{\top}(\epsilon \odot r_+) ~-~ t^{\top}\rho_0 ~-~ \zeta~,\\ 
& = \displaystyle \min_{\epsilon \in \mathcal{E}_{k}}~\max_{\substack{t \in \mathcal{T}_{F, c}}}\max_{\zeta \in \mathbb{R}}  ~~~~  \dfrac{-1}{2\lambda} \epsilon^{\top} \left(r_+ \odot \left(2r - r_+\right) \right) - t^{\top} \rho_0 - \zeta~,\\
\end{eqnarray*}
which can be written as
\begin{eqnarray*}
& \displaystyle \min_{\epsilon \in  \mathcal{E}_{k}}~\max_{\substack{t \in \mathcal{T}_{F, c}}}\max_{\zeta \in \mathbb{R}}  ~~~~  - \dfrac{1}{2\lambda} \sum_{v: r_v \geq 0} \epsilon_v r_v^2 - t^{\top} \rho_0 - \zeta~\\
= & \displaystyle \min_{\epsilon \in  \mathcal{E}_{k}}~\max_{\substack{t \in \mathcal{T}_{F, c}}}\max_{\zeta \in \mathbb{R}}  ~~~~  - \dfrac{1}{2\lambda} \sum_{v: t_v \leq -\zeta} \epsilon_v r_v^2 - t^{\top} \rho_0 - \zeta~\\
= & \displaystyle \min_{\epsilon \in  \mathcal{E}_{k}}~\max_{\substack{t \in \mathcal{T}_{F, c}}}\max_{\zeta \in \mathbb{R}}  ~~~~  - \dfrac{1}{2\lambda} \sum_{v: t_v \leq -\zeta} \epsilon_v (t_v + \zeta)^2 - t^{\top} \rho_0 - \zeta~\\
= & \displaystyle \min_{\epsilon \in  \mathcal{E}_{k}}~\max_{\substack{t \in \mathcal{T}_{F, c}}}\max_{\zeta \in \mathbb{R}}  ~~~~  - \dfrac{1}{2\lambda} \sum_{v: t_v \leq -\zeta} \left(\epsilon_v (t_v + \zeta)^2 ~+~ 2 \lambda t_v \rho_{0, v} \right) ~-~ \sum_{v: t_v > -\zeta} t_v \rho_{0,v} -  \zeta~.\\
\end{eqnarray*}

\end{proof}

\end{document}